\documentclass[a4paper,twocolumn]{article}

\usepackage{graphicx}
\usepackage{graphicx}
\usepackage{amssymb}
\usepackage{url}
\usepackage{float}
\usepackage{dsfont}
\usepackage{algorithm}
\usepackage{algpseudocode}
\usepackage{color}
\usepackage{multirow}
\usepackage{bbm}
\usepackage{bm}
\usepackage{mathtools}
\usepackage{hyperref}

\usepackage[table,xcdraw]{xcolor}

\DeclareMathOperator*{\argmin}{argmin}

\def\mbf{\mathbf}

\def\L{{\cal L}}
\def\R {\mathbb{R}}
\date{}
\title{Transformer VQ-VAE for Unsupervised Unit Discovery and Speech Synthesis: ZeroSpeech 2020 Challenge}
\author{Andros Tjandra$^1$, Sakriani Sakti$^{1,2}$, Satoshi Nakamura$^{1,2}$ \\
\small{$^1$Nara Institute of Science and Technology, Japan}\\
\small{$^2$RIKEN, Center for Advanced Intelligence Project AIP, Japan}\\ \\
\texttt{\{andros.tjandra.ai6, ssakti, s-nakamura\}@is.naist.jp}}

\begin{document}

\maketitle
\begin{abstract}
	
In this paper, we report our submitted system for the ZeroSpeech 2020 challenge on Track 2019. The main theme in this challenge is to build a speech synthesizer without any textual information or phonetic labels. In order to tackle those challenges, we build a system that must address two major components such as 1) given speech audio, extract subword units in an unsupervised way and 2) re-synthesize the audio from novel speakers. The system also needs to balance the codebook performance between the ABX error rate and the bitrate compression rate. Our main contribution here is we proposed Transformer-based VQ-VAE for unsupervised unit discovery and Transformer-based inverter for the speech synthesis given the extracted codebook. Additionally, we also explored several regularization methods to improve performance even further.
	
\end{abstract}

\section{Introduction}
Many speech technologies such as automatic speech recognition (ASR) and speech synthesis (TTS) has been used widely around the world. However, most such systems only cover rich-resource languages. For low-resource languages, using such technologies remains limited since most of those systems require many labeled datasets to achieve good performance. Furthermore, some languages have more extreme limitations, including no written form or textual interpretation for their context. In the ZeroSpeech challenge \cite{dunbar2019zs}, we addressed the latter problem by only learning the language elements directly from untranscribed speech.

ZeroSpeech 2020 has three challenges: 2017 Track 1, 2017 Track 2, and 2019. In this paper, we focus on the ZeroSpeech 2019 Track, where the task is called as TTS without Text.  This challenge has two major components that need to be addressed: 1) given speech audio, extract subword-like units that contain only context information with an unsupervised learning method and 2) re-synthesize the speech into a different speaker's voice.

To simultaneously solve these challenges, our strategy is to develop a primary model that disentangles the speech signals into two major factors: context and speaker information. After the disentanglement process, a secondary model predicts the target speech representation given the context extracted by the primary model. Following our success in a previous challenge \cite{tjandra2019vqvae} in ZeroSpeech 2019 and several prior publications on quantization-based approaches \cite{chorowski2019unsupervised, baevski2019vq, eloff2019unsupervised}, we used a vector quantized variational autoencoder (VQ-VAE) as the primary model and a codebook inverter as the secondary model. To improve our result, we introduced Transformer module \cite{vaswani2017attention} to capture the long-term information from the sequential data inside the VQ-VAE. We also explored several regularization methods to improve the robustness and codebook discrimination scores. Based on our experiment, all these combined methods significantly improved the ABX error rate.

\section{Self-Attention and Transformer Module}
The Transformer is a variant of deep learning modules that consist of several non-linear projections and a self-attention mechanism \cite{vaswani2017attention}.
Unlike recurrent neural networks (RNNs) such as a simple RNN or long-short term memory (LSTM) \cite{hochreiter1997long} modules, a Transformer module doesn't have any recurrent connection between the previous and current time-steps. However, a Transformer module utilizes self-attention modules to model the dependency across different time-steps.

\begin{figure}[h]
	\caption{A Transformer module}\label{fig:transformer_block}
	\centering
	\includegraphics[width=0.33\linewidth]{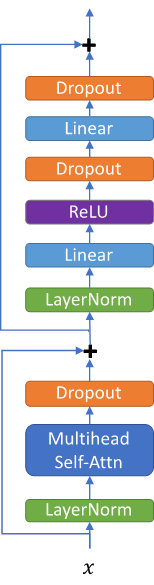}
	\vspace{-0.3cm}
\end{figure}

Given input sequence $X=[x_1, x_2, ..., x_S] \in \mathbb{R}^{S \times d_{in}}$ where $S$ denotes the input length and $d_{in}$ is the input dimension, a Transformer module produces hidden representation $Z=[z_1, z_2, ..., z_S] \in \mathbb{R}^{S \times d_{in}}$. Figure \ref{fig:transformer_block} shows the complete process inside a Transformer module.

Input sequence $X$ is normalized by a layer-norm and processed by a multi-head self-attention module. Self-attention module inputs are denoted as $Q$ (query), $K$ (key), and $V$ (value) where $\{Q,K,V\} \in \mathbb{R}^{S \times d_{k}}$. To calculate the output from a self-attention block, we apply the following formula:
\begin{align}
\texttt{Attention}(Q, K, V) = \texttt{softmax}\left(\frac{QK^{T}}{\sqrt{d_k}}\right) V.
\end{align}
Since a Transformer module employed multi-head self-attention instead of single head self-attention, input $Q, K, V$ needed to be varied to generate non-identical representation for each head. Therefore, for each head $p \in \{1..P\}$, we projected input $X$ with different matrices $W^Q_p, W^K_p, W^V_p$ and combined the output of many single head self-attentions by concatenations and a linear projection:
\begin{align}
\texttt{Multi-Head}(Q, K, V) = \texttt{Concat}(h_1, ..., h_P) W^O \\
\forall p \in \{1..P\}, h_p = \texttt{Attention}(QW_p^Q, KW_p^K, VW_p^V)
\end{align} where input projection matrices $W^Q_p \in \mathbb{R}^{d_{mdl} \times d_k}, W^K_p \in \mathbb{R}^{d_{mdl} \times d_k}, W^V_p \in \mathbb{R}^{d_{mdl} \times d_k}$, output projection matrix $W^O \in \mathbb{R}^{Pd_k \times d_{mdl}}$ and dimension of $d_k = \frac{d_{mdl}}{P}$. After we got the output from the self-attention, we applied layer normalization \cite{ba2016layer}, two linear projections, and a rectified linear unit (ReLU) activation function.

\section{Unsupervised Subword Discovery}
\label{sec:vqvae}
Normally, a speech utterance can be factored into several bits of latent information, including context, speaker's speaking style, background noise, emotions, etc. Here, we assume that the speech only contains two factors: context and the speaker's speaking style. In this case, the context denotes the unit that captures the speech information itself in the discretized form, which resembles phonemes or subwords. Therefore, to capture the context without any supervision, we used a generative model called a vector quantized variational autoencoder (VQ-VAE) \cite{van2017neural} to extract the discrete symbols. There are some differences between a VQ-VAE with a normal autoencoder \cite{vincent2008extracting} and a normal variational autoencoder (VAE) \cite{kingma2014adam} itself. The VQ-VAE encoder maps the input features into a finite set of discrete latent variables, and the standard autoencoder or VAE encoder maps input features into continuous latent variables. Therefore, a VQ-VAE encoder has stricter constraints due to the limited number of codebooks, which enforces the latent variable compression explicitly via quantization. On the other hand, standard VAE encoder has a one-to-one mapping between the input and latent variables. Due to the nature of the unsupervised subword discovery task, a VQ-VAE is more suitable for the subword discovery task compared to a normal autoencoder or VAE.

\begin{figure}[]
	\centering
	\includegraphics[width=0.8\linewidth]{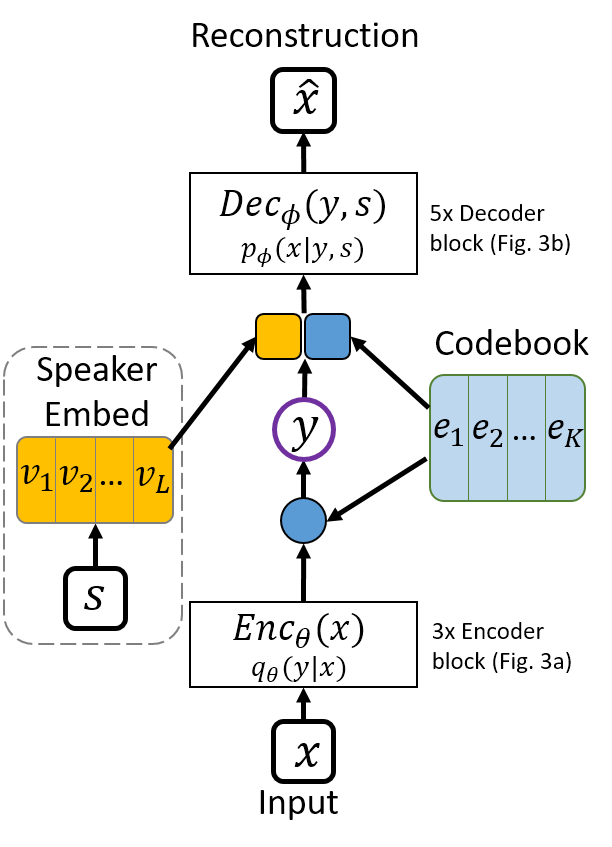}
	\caption{VQ-VAE for unsupervised unit discovery consists of several parts: encoder $\text{Enc}^{VQ}_\theta(x) = q_\theta(y|x)$, decoder $\text{Dec}^{VQ}_\phi(y, s) = p_\phi(x|y,s)$, codebooks $\mbf E =[e_1,..,e_K]$, and speaker embedding $\mbf V = [v_1,..,v_L]$.}
	\label{fig:vqvae}
\end{figure}

\begin{figure}[]
	\centering
	\includegraphics[width=0.8\linewidth]{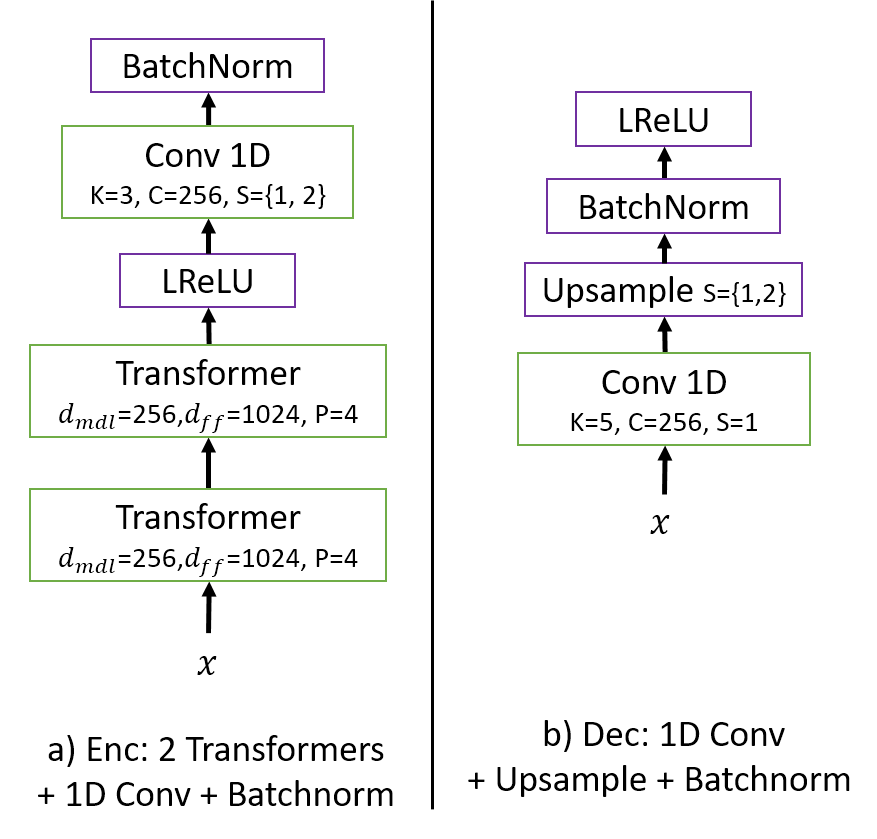}
	\caption{Building block inside VQ-VAE encoder and decoder: a) Encoder consisted of 2x Transformer layer and 1D convolution with stride to downsample input sequence length; b) Decoder consisted of 1D convolution, followed by up-sampling to recover original input sequence shape.}
	\label{fig:vqvae_resblock}
\end{figure}

We shows the VQ-VAE model in Fig.~\ref{fig:vqvae}. Here, we define $\mathbf{E} = [e_1,..,e_K] \in \R^{K \times D_e}$ as a collection of codebook vectors and $\mathbf{V} = [v_1,..,v_L] \in \R^{L \times D_v}$ as a collection of speaker embedding vectors. At the encoding step, input $x$ denotes such speech features as MFCC (Mel-frequency cepstral coefficients) or Mel-spectrogram and auxiliary input $s \in \{1,..,L\}$ denotes the speaker ID from speech feature $x$. In Fig.~\ref{fig:vqvae_resblock}, we show the details for the building block inside the encoder and decoder modules. Encoder $q_\theta(y|x)$ outputs a discrete latent variable $y \in \{1,..K\}$. To transform an input in continuous variable into a discrete variable, the encoder first produces intermediate continuous representation ${z} \in \R^{D_e}$. Later on, we scan through all codebooks to find which codebook has a minimum distance between ${z}$ and a vector in $\mathbf{E}$. We define this process by the following equations:
\begin{align}
q_{\theta}(y=c|x) &= \begin{cases}
1 \quad \text{if } \, c=\argmin_{i} \text{Dist}( {z}, e_i )\\
0 \quad \text{else }
\end{cases} \\
e_c &= \mathbb{E}_{q_\theta (y|x)} [\mbf E] \\ &= \sum_{i=1}^{K} q_\theta(y=i | x) \, e_i.
\end{align} where $\text{Dist}(\cdot, \cdot): \R^{D_e} \times \R^{D_e} \rightarrow \R$ is a function that calculates the distance between two vectors. In this paper, we define $\text{Dist}(a, b) = \| a - b \|_2$ as the L2-norm distance.

After the closest codebook index $c \in \{1,..,K\}$ is found, we substitute latent variable ${z}$ with nearest codebook vector $e_c$. Later, decoder $p_\phi(x|y, s)$ uses codebook vector $e_c$ and speaker embedding $v_s$ to reconstruct input feature $\hat{x}$.

\subsection{VQ-VAE objective}
Given input reconstruction across all time-step $\forall t \in [1..T]$, $\hat{X} = [\hat{x}_1,..,\hat{x}_T]$ and closest codebook index $[c_1,..,c_T]$, we calculate the VQ-VAE objective:
\begin{align}
\mathcal{L}_{VQ} = \sum_{t=1}^{T} -\log p_\phi(x_t|y_t, s) + \gamma \| {z_t} - \text{sg}(e_{c_{t}}) \|_2^2, \label{eq:loss_vq},
\end{align} where function $\text{sg}(\cdot)$ stops the gradient:
\begin{align}
x = sg(x) ; \quad \quad
\frac{\partial \,\text{sg}(x)}{\partial \,x} = 0.
\end{align}
The first term is a negative log-likelihood to measure the reconstruction loss between original input $x_t$ and reconstruction $\hat{x}_t$ to optimize encoder parameters $\theta$ and decoder parameters $\phi$. The second term minimizes the distance between intermediate representation ${z_t}$ and nearest codebook $e_{c_{t}}$, but the gradient is only back-propagated into encoder parameters $\theta$ as commitment loss. The impact from commitment loss controlled with a hyper-parameter $\gamma$ (we use $\gamma=0.25$). To update the codebook vectors, we use an exponential moving average (EMA) \cite{kaiser2018fast}. EMA updates the codebook $\mbf E$ independently regardless of the optimizer's type and update rules, therefore the model is more robust against different optimizer's choice and hyper-parameters (e.g., learning rate, momentum) and also avoids posterior collapse problem \cite{roy2018towards}.

\subsection{Model regularization}
\subsubsection{Temporal smoothing}
Since our input datasets are sequential data, we introduced temporal smoothing between the encoded hidden vectors between two consecutive time-steps:
\begin{align}
\mathcal{L}_{reg} = \sum_{i=1}^{T-1}\|z_t - z_{t+1}\|_2^2.
\end{align}
The final loss is defined:
\begin{align}
\mathcal{L} = \mathcal{L}_{VQ} + \lambda \mathcal{L}_{reg},
\end{align} where $\lambda$ denotes the coefficient for the regularization term.
\subsubsection{Temporal jitter}
Temporal jitter regularization \cite{chorowski2019unsupervised} is used to prevent the latent vector co-adaptation and to reduce the model sensitivity near the unit boundary. In the practice, we could apply the temporal jitter by:
\begin{align}
j_t &\sim Categorical(p, p, 1-2*p) \in \{1,2,3\}\\
\hat{c_t} &= \begin{cases}
{c_{t-1}}, & \text{if } j_t = 1 \text{ and } t>1\\
{c_{t+1}}, & \text{if } j_t = 2 \text{ and } t<T\\
{c_t}, & \text{else } 
\end{cases} \\
e_{t} &= \mathbf{E}[{\hat{c_t}]}
\end{align} where $p$ is the jitter probability, $c_t$ is the closest codebook index at time-$t$ and $\hat{c_t}$ is the new assigned codebook index after the jitter operation.


\section{Codebook Inverter}
\label{sec:inverter}
A codebook inverter model is used to generate the speech representation given the predicted codebook from our Transformer VQ-VAE. The input is $[\mbf E[c_1],..,\mbf E[c_T]]$, and the output is the following speech representation sequence (here we use linear magnitude spectrogram): $\mbf X^{R} = [\mbf X^R_1, .., \mbf X^R_{S}]$.

\begin{figure}[h]
	\centering
	\includegraphics[width=0.8\linewidth]{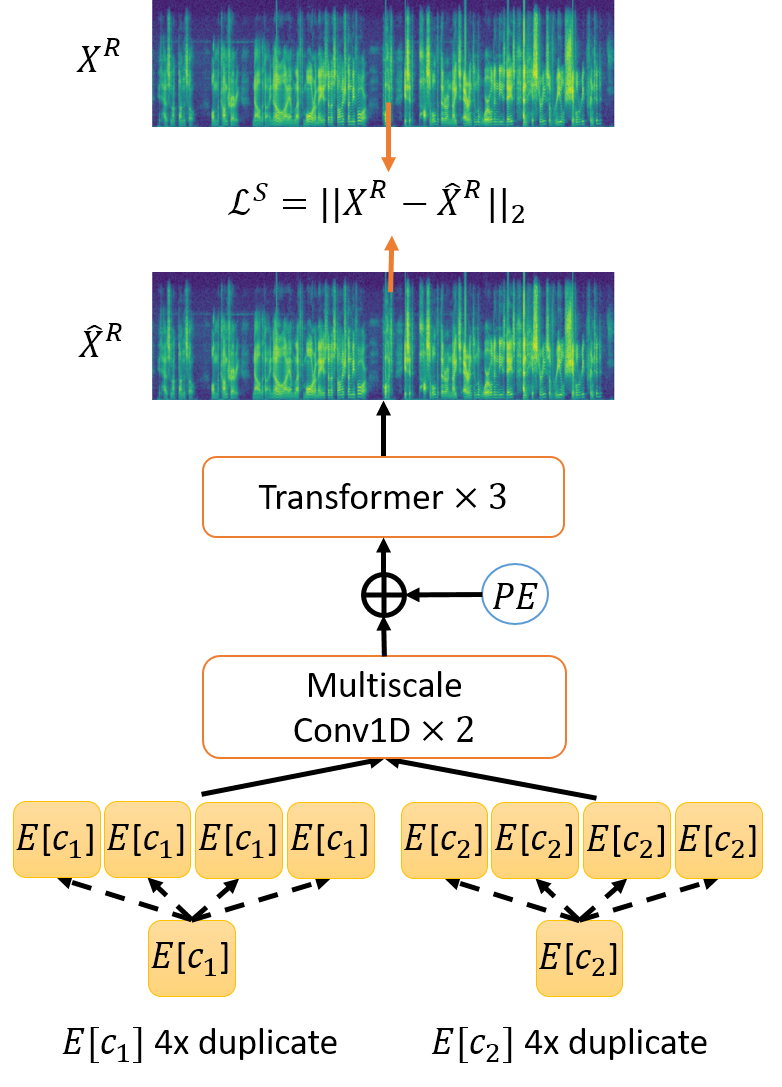}
	\caption{Codebook inverter: given codebook sequence $[\mbf E[y_1], .., \mbf  E[y_{T}]]$, we predict corresponding linear magnitude spectrogram $\hat{\mbf X}^R = [x^R_1, .. x^R_{S}]$. If lengths between $S$ and $T$ are different, we consecutively duplicate each codebook by $r$-times.}
	\label{fig:codeinverter}
\end{figure}

In Fig.~\ref{fig:codeinverter}, we show our codebook inverter architecture that consists of two multiscale 1D convolutions and three Transformer layers with additional sinusoidal position encoding to help the model distinguish the duplicated codebook positions.
Depends on the VQ-VAE encoder architecture, the predicted codebook sequence length $T$ could be shorter than $S$ because the VQ-VAE encoder $q_\theta(y|x)$ have several convolutional layers with stride larger than one. Therefore, to re-align the codebook sequence representation with the speech representation, we duplicate each codebook occurrences $\forall t \in \{1..T\}\, \mathbf{E}[{c_t}]$ into $r$ copies side-by-side where $r= S/T$.
To train a codebook inverter, we set the objective function:
\begin{eqnarray}
\L_{INV} =  \| \mbf X^R - \hat{\mbf X}^R \|_{2} \label{eq:loss_inv}
\end{eqnarray} to minimize the L2-norm between predicted spectrogram $\hat{\mbf X}^R = \text{Inv}_{\rho}([\mathbf{E}[c_1], ..., \mathbf{E}[c_T]])$ and groundtruth spectrogram $\mbf X^R$. We defined $\text{Inv}_{\rho}$ as the inverter parameterized by $\rho$. During the inference step, Griffin-Lim \cite{griffin1984signal} is used to reconstruct the phase from the spectrogram and applied an inverse short-term Fourier transform (STFT) to invert it into a speech waveform.
\section{Experiment}
In this section, we describe our pipeline, including feature preprocessing, model settings, and hyperparameters.

\subsection{Experimental Set-up}
There are two datasets for two languages, English data for the development dataset, and a surprising  Austronesian language for the test dataset.
Each language dataset contains subset datasets: (1) a voice dataset for speech synthesis, (2) a unit discovery dataset, (3) an optional parallel dataset from the target voice to another speaker's voice, and (4) a test dataset. The source corpora for the surprise language are described here \cite{indotts, indoasr}, and further details can be found \cite{zs2019}. We only used (1) and (2) for training the VQ-VAE codebook and inverter and (4) for evaluation.

For the speech input, we experimented with two different feature representations, such as Mel-spectrogram (with 80 dimensions, 25-ms window size, 10-ms time-steps) and MFCC (with 39 dimensions, 13 main $+\Delta+\Delta^2$, 25-ms window size, 10-ms time-steps). Both MFCC and Mel-spectrogram are generated by the Librosa package \cite{mcfee2015librosa}. All models are implemented with PyTorch library \cite{NEURIPS2019_9015}. We use Adam \cite{kingma2014adam} optimizer with learning rate $1e-4$ for training both models.

\subsection{Experimental Results}
In this subsection, we report our experiment based on different scenarios and hyperparameters with the English dataset.
First, we use different features as the input for VQ-VAE in Table \ref{tab:result_vqvae_diff_feat}. We compared the effect of using log Mel-spectrogram with 80 dimensions versus MFCC$+ \Delta + \Delta^2$ with 39 dimensions.

\begin{table}[h]
	\centering
	\caption{ABX and Bitrate result between MFCC and log Melspectrogram features on Transformer VQ-VAE (TrfVQVAE with $K=128$, stride $4\times$)}
	\label{tab:result_vqvae_diff_feat}
	\begin{tabular}{|l|l|l|}
		\hline
		\multicolumn{1}{|c|}{\textbf{Model}} & \textbf{ABX} & \textbf{Bitrate} \\ \hline
		TrfVQVAE with log-Mel & 33.79 & 171.05 \\ \hline
		TrfVQVAE with MFCC & 21.91 & 170.42 \\ \hline
	\end{tabular}
\end{table}

Based on the result on Table~\ref{tab:result_vqvae_diff_feat}, we determined that MFCC provides much better results in terms of ABX error rate. From this point, we use MFCC as the VQ-VAE input features. In Table \ref{tab:result_vqvae_diff_k_t}, we compared our proposed model with our submission from previous challenge \cite{tjandra2019vqvae} and we also explored different codebook sizes to determine the trade-off between the ABX error rate and bitrate.
\begin{table}[h]
	\centering
	\small
	\caption{ABX and Bitrate result between our last year submission and new proposed Transformer VQ-VAE.}
	\label{tab:result_vqvae_diff_k_t}
	\begin{tabular}{|c|l|l|}
		\hline
		\multicolumn{1}{|c|}{\textbf{Model}} & \textbf{ABX} & \textbf{Bitrate} \\ \hline
		Conv VQVAE (stride $4\times$, K=256) \cite{tjandra2019vqvae} & 24.17 & 184.32 \\ \hline
		TrfVQVAE (stride $4\times$, K=64) \, & 22.72 & 141.82 \\ \hline
		TrfVQVAE (stride $4\times$, K=128) & 21.91 & 170.42 \\ \hline
		TrfVQVAE (stride $4\times$, K=256) & 21.94 & 194.69 \\ \hline
		TrfVQVAE (stride $4\times$, K=512) & 21.6 & 217.47 \\ \hline
	\end{tabular}
\end{table}

Based on the result on Table~\ref{tab:result_vqvae_diff_k_t}, by using Transformer VQ-VAE, we outperformed our previous best submission by -2.2 ABX error rate and with a similar bitrate. Later on, by using small codebook sizes $K=64$, we reduce the bitrate by -30 points compared to $K=128$, but we sacrifice the +0.8 ABX error rate. 
We explored different regularization methods to improve transformer VQ-VAE’s performance in Table \ref{tab:result_vqvae_temp_smooth}. In Table \ref{tab:result_vqvae_temp_smooth}, we show the result from Transformer VQ-VAE (stride $4\times$, K=128) with different $\lambda$ smoothing coefficient, different jitter probability and the combination between both regularization methods. We found out that by combining both temporal smoothing and jittering, we get -1.77 ABX error rate improvement.
\begin{table}[h]
	\centering
	\small
	\caption{ABX and Bitrate result between different temporal smoothing coefficient $\lambda$, jitter probability $p$, and their combination for Transformer VQ-VAE regularization. (Notes: $\bigstar$ denotes our submitted system. After the submission, we continue the experiment and we found out another hyperparameters combination brings even lower ABX than our submitted system.)}
	\label{tab:result_vqvae_temp_smooth}
	\begin{tabular}{|c|l|l|}
		\hline
		\multicolumn{1}{|c|}{\textbf{Model}} & \textbf{ABX} & \textbf{Bitrate} \\ \hline
		TrfVQVAE (stride $4\times$, K=128) & 21.91 & 170.42 \\ \hline 
		+ temp smooth $(\lambda=1e-2)$ & 21.88 & 169.02 \\ \hline
		+ temp smooth $(\lambda=5e-3)$ & 21.67 & 169.2 \\ \hline
		+ temp smooth $(\lambda=1e-3)$ & 21.75 & 169.56 \\ \hline \hline
		+ temp jitter $(p=0.05)$ \, & 21.57 & 166.19 \\ \hline
		+ temp jitter $(p=0.075)$ & 21.70 & 164.08 \\ \hline \hline
		\begin{tabular}{@{}c@{}}+ temp smooth $(\lambda=5e-3)$ $\bigstar$  \\+ temp jitter $(p=0.05)$ \end{tabular}& {20.71} & {171.99} \\ \hline
		\begin{tabular}{@{}c@{}}+ temp smooth $(\lambda=1e-3)$ \\+ temp jitter $(p=0.05)$ \end{tabular}& {20.14} & {167.02}
		\\ \hline
	\end{tabular}
	\vspace{-0.2cm}
\end{table}

\section{Conclusions}
We described our approach for the ZeroSpeech 2020 challenge on Track 2019. For the unsupervised unit discovery task, we proposed a new architecture: Transformer VQ-VAE to capture the context of the speech into a sequence of discrete latent variables. Additionally, we also use the Transformer block inside our codebook inverter architecture. Compared to our last year's submission, replacing convolutional layers with Transformer gives a significant improvement up to -2.2 ABX error rate. To improve the result further, we investigated several regularization methods. In the end, combining both temporal smoothing and jittering improved the Transformer VQ-VAE performance up to -1.77 ABX error rate compared to the un-regularized model.

\section{Acknowledgements}

Part of this work was supported by JSPS KAKENHI Grant Numbers JP17H06101 and JP17K00237.

\bibliographystyle{IEEEtran}

\bibliography{mybib}


\end{document}